\documentclass{article}

\PassOptionsToPackage{numbers, compress}{natbib}

\usepackage[preprint]{neurips_2024}

\usepackage[utf8]{inputenc} 
\usepackage[T1]{fontenc}    
\usepackage{hyperref}       
\usepackage{url}            
\usepackage{booktabs}       
\usepackage{amsfonts}       
\usepackage{nicefrac}       
\usepackage{microtype}      
\usepackage[table,usenames,dvipsnames]{xcolor}         

\usepackage{comment}
\usepackage{listings}
\usepackage{tikz}
\newcommand{\exmy}{eXmY}
\newcommand{\keyword}[1]{\texttt{\textcolor{ForestGreen}{#1}}}

\title{eXmY: A Data Type and Technique for \\
Arbitrary Bit Precision Quantization}

\author{%
  Aditya Agrawal \\
  \texttt{adityaag@google.com} \\
  \And
  Matthew Hedlund \\
  \And
  Blake Hechtman \\
  \texttt{blakehechtman@google.com}
  \AND
  Google LLC
}

\begin{document}

\maketitle

\begin{abstract}
\exmy\ is a novel data type for quantization of ML models.
It supports both arbitrary bit widths and arbitrary
integer and floating point formats.
For example, it seamlessly supports 3, 5, 6, 7, 9 bit formats.
For a specific bit width, say 7, it defines all possible
formats e.g. e0m6, e1m5, e2m4, e3m3, e4m2, e5m1 and e6m0.
For non-power of two bit widths e.g. 5, 6, 7, we created a
novel encoding and decoding scheme which achieves perfect
compression, byte addressability and is amenable to
sharding and vector processing.
We implemented libraries for emulation, encoding
and decoding tensors and checkpoints in C++, TensorFlow,
JAX and PAX.
For optimal performance, the codecs use SIMD instructions 
on CPUs and vector instructions on TPUs and GPUs.
\exmy\ is also a technique and exploits the statistical
distribution of exponents in tensors.
It can be used to quantize weights, static and 
dynamic activations, gradients, master weights and
optimizer state.
It can reduce memory (CPU DRAM and accelerator HBM),
network and disk storage and transfers.
It can increase multi tenancy and accelerate compute.
\exmy\ has been deployed in production for almost 2 years.
\end{abstract}

\section{Introduction}

The relentless growth in model size poses significant challenges
for model training, pretraining, finetuing and serving. 
Large Embedding Models (LEMs) e.g. DLRM \cite{model:dlrm}
and Large Language Models (LLMs) e.g.
PaLM \cite{model:palm},
LLaMA \cite{model:llama, model:llama2, model:llama3},
GPT-3 \cite{model:gpt3},
have large memory footprint, memory and network bandwidth requirements,
compute requirements, serving latencies, energy consumption and cost.

Quantization is a proven approach to mitigate these challenges,
by reducing the precision of model weights, master weights,
activations, gradients, optimizer states, and network communication.
However, most existing quantization techniques and hardware rely on
conventional power-of-two bit widths and formats, which may not be
ideally suited for preserving model quality in all use cases.

Previously, ML accelerators e.g. Google TPUs
\cite{hardware:tpu_v4, hardware:tpu_v5e, hardware:tpu_v5p},
and Nvidia GPUs \cite{hardware:nv_ampere, hardware:nv_hopper}
added support for \keyword{int8} and \keyword{int4} datatypes.
More recently, Nvidia H100 \cite{hardware:nv_hopper} and
Nvidia GB200 \cite{hardware:nv_blackwell}
have added support for \keyword{fp8}, \keyword{fp6}
and \keyword{fp4} datatypes.
Nvidia TensorFloat32 \cite{format:tf32} and the
OCP \cite{format:ocpmx} \keyword{fp6} formats e.g.
\keyword{e2m3} and \keyword{e3m2} are a step in the direction
of supporting non power-of-two bit widths.
However, they do not address the entire problem space.
In addition, they do not provide a bit packing and unpacking
scheme to actually reduce the memory footprint and bandwidth.

Different layers and operations within a model have different sensitivity to precision, for example, the authors in \cite{format:nv_fp8} suggest
using \keyword{e4m3} for weight and activation tensors, and
\keyword{e5m2} for gradient tensors.
In this work, we propose and advocate the use of flexible,
arbitrary bit precision formats which can be tailored to the
specific requirements of each model component e.g.
master weights, training weights, serving weights,
network communication etc.
Our contributions are
\begin{itemize}
\item A novel datatype which supports arbitrary bit widths and formats.
\item A software library to emulate any datatype using existing
\keyword{bfloat16} or \keyword{float32} datatypes.
This enables very fast evaluation of model quality at different
formats and bit widths.
The library can be used to quantize weights, master weights,
static and dynamic activations, gradients, optimizer states
and network communication.
The library preserves \keyword{NaNs} and \keyword{Infs} for
easy debugging.
\item Software codecs for packing and unpacking bits into existing
datatypes. The codecs achieve perfect compression, offer byte
addressability, works seamlessly with sharding and are amenable
to vector processing on CPUs, GPUs and TPUs for high performance.
\item Discovered a distribution of exponents in ML models and
proposed a technique to exploit the distribution to significantly
reduce the number of bits required by ML models.
\end{itemize}

\section{A New Datatype}
\begin{table}[b]
    \caption{Floating point datatypes.}
    \centering
    \renewcommand{\arraystretch}{1.5}
    \rowcolors{1}{RoyalBlue!10}{}
    \begin{tabular}{|c|c|c|c|c|c|c|}
    \toprule \hline
    \rowcolor{RoyalBlue!40}
    Format & AKA & \# Bits & Sign Bit & \# Exponent Bits & \# Mantissa Bits & Exponent Bias \\
    fp32 & e8m23 & 32 & 1 & 8 & 23 & 127 \\
    tf32 & e8m10 & 19 & 1 & 8 & 10 & 127 \\
    bf16 & e8m7 & 16 & 1 & 8 & 7 & 127 \\
    fp16 & e5m10 & 16 & 1 & 5 & 10 & 15 \\
    fp8 & e4m3 & 8 & 1 & 4 & 3 & 7 \\
    fp8 & e5m2 & 8 & 1 & 5 & 2 & 15 \\
    {} & \exmy\ & $1 + X + Y$ & 1 & $X$ & $Y$ & variable \\
    \hline \bottomrule
    \end{tabular}
    \label{table:datatype}
\end{table}

Over the years, many floating point formats have been
proposed. Some of those have been IEEE standardized e.g.
\keyword{float64}, \keyword{float32} and \keyword{float16}
\cite{format:ieee754}. Some are vendor specific e.g.
\keyword{bfloat16} from Google \cite{format:bf16} and
\keyword{tensorfloat32} from NVidia \cite{format:tf32}.
Others like \keyword{fp8}, \keyword{fp6}, \keyword{fp4}
\cite{format:ocpmx} have been proposed recently by the
Open Compute Project (OCP).
Some formats like \keyword{float32} have only one definition
i.e., 1 sign bit, 8 exponent bits, 23 mantissa bits,
exponent bias of $127$, supports subnormals, NaNs,
positive and negative infinities,
while, others like \keyword{fp8} support multiple formats
within the same bit width e.g. \keyword{e4m3} and
\keyword{e5m2}.
Table \ref{table:datatype}, shows the bit allocation and
exponent bias for a few different data types.

\begin{description}

\item[Format]
\exmy\ is a generalization of the floating point format to
arbitrary bit widths and formats. It has 1 sign bit,
$X$ exponent bits and $Y$ mantissa bits.
For example, with 7 bits, it defines 7 formats viz. \keyword{e6m0},
\keyword{e5m1}, \keyword{e4m2}, \keyword{e3m3}, \keyword{e2m4},
\keyword{e1m5} and \keyword{e0m6}.

When $X = 1$, the format becomes linear and equivalent to a 
symmetric signed integer format, e.g.
\keyword{e1m2} is equivalent to symmetric \keyword{int4}
and can represent integers from $[-7, 7]$,
\keyword{e1m3} is equivalent to symmetric \keyword{int5}
and can represent integers from $[-15, 15]$ etc.
This equivalence enables comparing integer and floating
point formats more easily, for example, their dynamic range
and precision.
It also enables implementing integer arithmetic using
floating point hardware.

When $X = 0$, the format degenerates to the form (sign, magnitude).
Like floating point numbers, it has a double zero, but
it can be instead interpreted as a 2's complement number to get an
additional encoding. For example, \keyword{e0m3} can be used as
\keyword{int4} and represent integers from $[-8, 7]$.

Therefore, \exmy\ can represent signed integers, symmetric
signed integers and floating point numbers.
Overall, for bit widths less than and equal to 8, it defines
36 different formats from \keyword{e7m0} down to \keyword{e0m0}.
For bit widths between 8 and 32 there are dozens of
formats e.g. \keyword{e5m4}.

\item[Subnormals]
Subnormals, i.e. an exponent value of zero and non zero
mantissa, increase the dynamic range of the representation.
\exmy\ supports subnormals like other floating point formats.

\item[Rounding]
The IEEE 754 standard \cite{format:ieee754} defines 5
rounding modes viz. \keyword{roundTiesToEven},
\keyword{roundTiesToAway}, \keyword{roundTowardPositive},
\keyword{roundTowardNegative} and \keyword{roundTowardZero}.
The rounding mode  \keyword{roundTiesToEven}, also referred to as,
Round To Nearest Even (RTNE), is the default rounding mode for
binary formats.
We extended the RTNE logic in Eigen \cite{software:eigen}
for rounding from \keyword{float32} to \keyword{bfloat16}, to
arbitrary number of mantissa bits.
We preserve NaNs and Infs during rounding.

\item[NaNs \& Infs]
Support for \keyword{NaNs} and \keyword{Infs} is optional in \exmy.
This is especially important for serving in sub byte precision,
because trained ML model weights do not have \keyword{NaNs}
or \keyword{Infs}.

\item[Exponent Bias]
In the IEEE and OCP formats, the exponent bias, the smallest normal
and the normal exponent range are defined by the standard.
These values are interdependent and there is only 1 degree of freedom.
For example, in the IEEE \keyword{float32} format, the exponent bias
is $127$, the smallest normal is $2^{-126}$ and the normal
exponent range is $[2^{-126}, 2^{127}]$.
For the OCP \keyword{E4M3} format, the corresponding values
are $7$, $2^{-6}$ and $[2^{-6}, 2^{8}]$.
However, in \exmy, these values are software defined
and is stored as metadata. For example, in \keyword{e3m3},
with 3 exponent bits, the corresponding values could be
($2$, $2^{-1}$, $[2^{-1}, 2^5]$) or ($-1$, $2^2$, $[2^2, 2^8]$)
i.e. the value $2^0 = 1$ is not even in the normal exponent
range in the second example.

\item[Metadata]
Since the byte and sub-byte formats have limited dynamic range,
\exmy, OCP formats \cite{format:ocpmx}, conventional \keyword{int8}
and \keyword{int4} quantization schemes, maintain some metadata.
Typically, with \keyword{int8} and \keyword{int4} quantization,
the metadata is a \keyword{bfloat16} or \keyword{float32}
scaling factor.
In the OCP formats, the metadata is an 8-bit power-of-2
scaling factor. Its format is the same as the 8-bit exponent
field of the IEEE \keyword{float32} format.
In \exmy, the metadata is the value of the maximum biased
exponent, an 8 bit value.

The maximum biased exponent can be determined before or after
rounding to the appropriate number of mantissa bits.
An additional \keyword{bfloat16} or \keyword{float32} scaling
factor can also be maintained.

\item[Block Size]
The OCP formats define a block size of $32$, i.e.
the metadata is shared between $32$ elements.
\exmy\ does not define or constrain the size or shape of
the block. A block can be a tensor, a row, a column, a sub row
or even a 2D tile. As is obvious, more metadata improves
model quality at the expense of storage.
In general, we have observed that for LLM serving,
\keyword{e3m2} and \keyword{e3m1} require only one
metadatum per row, while \keyword{e2m1} and \keyword{e1m2}
benefit from smaller block sizes.

\end{description}

\subsection{Emulation}

Just like we can emulate \keyword{int5} or \keyword{int7}
using an \keyword{int8} datatype,
likewise, we can emulate any \exmy\ format
using \keyword{bfloat16}, if $X \le 8$ and $Y \le 7$, or
using \keyword{fp16}, if $X \le 5$ and $Y \le 10$, or
using \keyword{float32}, if $X \le 8$ and $Y \le 23$.
We preserve NaNs and Infs during emulation.

\begin{figure}[h]
    \centering
    \includegraphics[width=\linewidth]{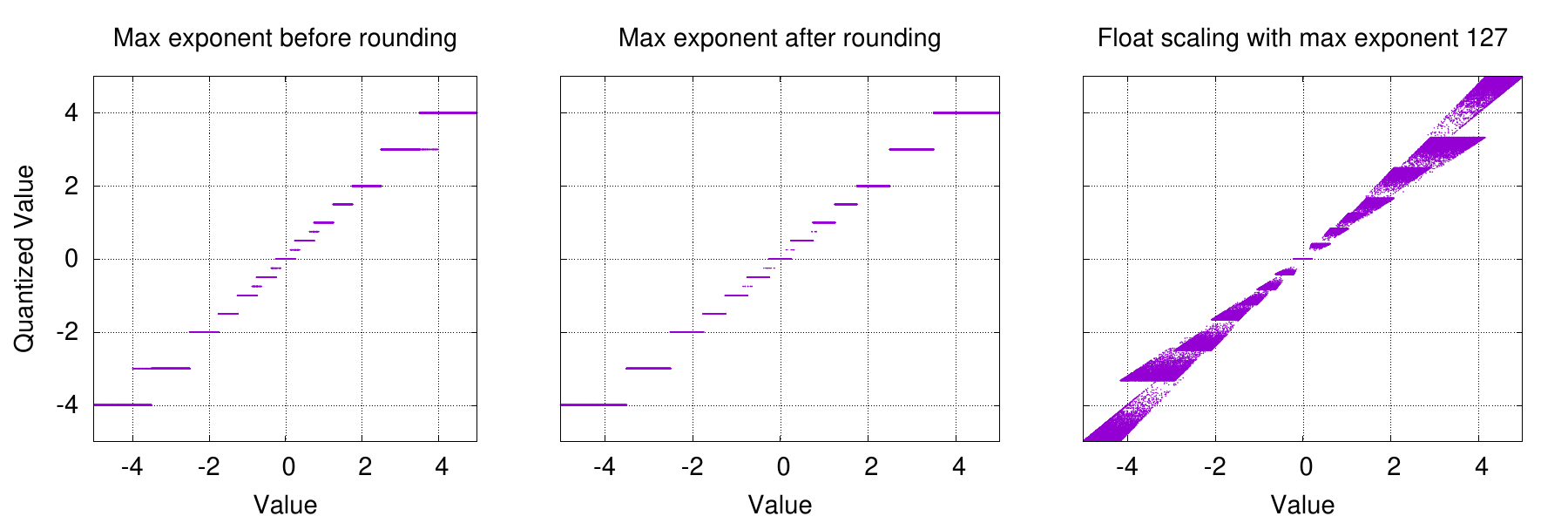}
    \caption{Emulation using \keyword{e2m1} with different schemes.}
    \label{fig:emulation}
\end{figure}

Fig. \ref{fig:emulation} shows a scatter plot of the
original values vs. values emulated with \keyword{e2m1}
at block size 16 and at three different schemes viz.
maximum exponent before rounding, maximum exponent after rounding,
and float scaling with maximum exponent of 127.
Note that the same input value can either be
(a) saturated to the largest normal,
(b) rounded to the appropriate number of mantissa bits,
(c) considered a subnormal, or
(d) flushed to zero,
depending on its relative value in the block.

The first two plots have a staircase pattern with one step 
between every power of two.
The scheme maximum exponent after rounding is useful at small
block sizes to prevent excessive truncation of the largest value
in the block.
For example, in the first scheme, 3.9 either rounds down to 3.0
or rounds up to 4.0,
while, in the second scheme, it always rounds up to 4.0.
The float scaling scheme captures the largest value in
the block accurately.

\subsection{Codecs: Encoder \& Decoder}

Current processors provide only a few compute data types
e.g. \keyword{float32}, \keyword{bfloat16}, \keyword{int8},
\keyword{int4}, OCP \keyword{e4m3} etc.,
however, \exmy\ supports dozens of formats. Therefore,
we need software routines or hardware instructions to
encode and decode from \exmy\ data types.

The encoding and decoding can be done offline or on the fly.
For example, trained weights and static activations (feature maps)
can be encoded offline and is not performance critical.
However, decoding weights during serving or
encoding and decoding the dynamic activations and gradients
before and after network communication is performance critical.
The codecs have two components:

\subsubsection{Type Conversion: Float $\longleftrightarrow$ \exmy\ + Metadata}
In this step, we convert a float format to an \exmy\
format and store it an 8, 16 or 32 bit container and vice versa.
The metadata i.e., maximum exponent, is maintained separately.
For example, we convert a \keyword{bfloat16} tensor of
shape $(R, C)$ to an \keyword{int8} tensor of shape
$(R, C)$ containing \keyword{e3m3} values,
and an \keyword{int8} tensor of shape $(R, 1)$ containing
the metadata.

\subsubsection{Bit Packing \& Unpacking: Power-of-2 Decomposition}

\begin{figure}[!h]
    \centering
    \begin{tikzpicture}[scale=\textwidth/12.5cm]

\pgfmathsetmacro{\bitw}{0.25}
\pgfmathsetmacro{\bith}{0.5}
\definecolor{color1}{RGB}{251,180,174}
\definecolor{color2}{RGB}{179,205,227}
\definecolor{color3}{RGB}{204,235,197}

\begin{scope}[shift={(2, 2)}, font=\small]
  \node at (4*\bitw, 8.5*\bith) {7 bits};
  \node [rotate=90] at (-0.25, 4*\bith) {8 elements};

  \foreach \i in {0} {
    \foreach \j in {0,1,...,7} {
      \draw [fill=white] ({\bitw * \i}, \bith * \j) rectangle ({\bitw * \i + \bitw}, {\bith * \j + \bith});
    }
  }

  \foreach \i in {1} {
    \foreach \j in {0,1,...,7} {
      \draw [fill=color3] ({\bitw * \i}, \bith * \j) rectangle ({\bitw * \i + \bitw}, {\bith * \j + \bith});
    }
  }
  \foreach \i in {2,3} {
    \foreach \j in {0,1,...,7} {
      \draw [fill=color2] ({\bitw * \i}, \bith * \j) rectangle ({\bitw * \i + \bitw}, {\bith * \j + \bith});
    }
  }
  \foreach \i in {4,5,6,7} {
    \foreach \j in {0,1,...,7} {
      \draw [fill=color1] ({\bitw * \i}, \bith * \j) rectangle ({\bitw * \i + \bitw}, {\bith * \j + \bith});
    }
  }
\end{scope}

\draw [<->, thick] (4.25,4.25) -- (5,4.25);

\begin{scope}[shift={(6, 5)}, font=\small]
  \node at (-0.5, 0.5*\bith) {int32};
  \foreach \i in {0,1,...,31} {
    \draw [fill=color1] ({\bitw * \i}, 0) rectangle ({\bitw * \i + \bitw}, {\bith});
  }
\end{scope}

\begin{scope}[shift={(6, 4)}, font=\small]
  \node at (-0.5, 0.5*\bith) {int16};
  \foreach \i in {0,1,...,15} {
    \draw [fill=color2] ({\bitw * \i}, 0) rectangle ({\bitw * \i + \bitw}, {\bith});
  }
\end{scope}

\begin{scope}[shift={(6, 3)}, font=\small]
  \node at (-.5, 0.5*\bith) {int8};
  \foreach \i in {0,1,...,7} {
    \draw [fill=color3] ({\bitw * \i}, 0) rectangle ({\bitw * \i + \bitw}, {\bith});
  }
\end{scope}

\end{tikzpicture}
    \caption{Bit packing and unpacking for 7-bit wide elements.}
    \label{fig:bit_packing_unpacking}
\end{figure}
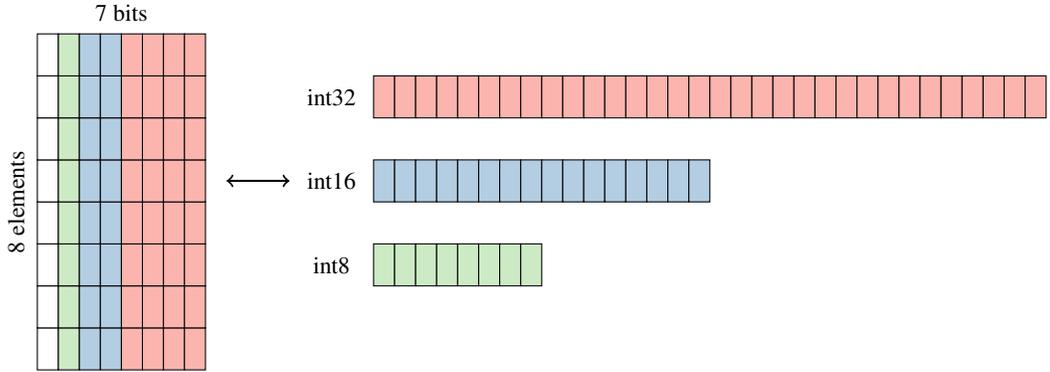

Consider an array, where each element is a 7-bit \exmy\
datatype e.g., \keyword{e3m3}.
Fig. \ref{fig:bit_packing_unpacking}, shows the scheme for
packing and unpacking an array of shape $(8, 1)$ with
7-bit elements.
Before packing and after unpacking, the elements are held in
an \keyword{int8} container as shown in the figure.
We decompose the bits into \emph{power-of-2} segments i.e.,
$7 = 4 + 2 + 1$. We pack
eight 4-bit elements into an \keyword{int32} container,
eight 2-bit elements into an \keyword{int16} container, and 
eight 1-bit elements into an \keyword{int8} container,
as shown on the right.
Overall, an array of shape $(8R, C)$ gets packed into 3
arrays of \keyword{int32}, \keyword{int16} and \keyword{int8},
each of shape $(R, C)$.
There are many advantages of this scheme.

\begin{itemize}
    \item{Uses existing storage datatypes e.g. \keyword{int32},
    \keyword{int16} and \keyword{int8}.}
    \item{Independent of the data format e.g., the 7-bit format
    could be either \keyword{e4m2} or \keyword{e3m3}.}
    \item {Achieves perfect compression i.e., 8 elements of
    7-bits use exactly $32 + 16 + 8 = 56$ bits.}
    \item{Works for all arbitrary bit widths.
    For example, an array of shape $(8R, C)$ containing 5-bit elements
    ($5 = 4 + 1$) can be packed into 2 arrays of \keyword{int32}
    and\keyword{int8}, each of shape $(R, C)$.}
    \item{Amenable to SIMD and vector processing on current CPUs, GPUs and TPUs.}
    \item{The array can be sharded along rows or columns, before
    or after packing, and each shard can be independently reconstructed.}
    \item{Can be modified to pack along columns i.e.,
    an array of shape $(R, 8C)$ can be packed into multiple arrays
    of shape $(R, C)$.}
    
\end{itemize}

The only constraint of the scheme is that the number of rows
or columns is a multiple of 8, which is almost always true
in ML models.

\section{Technique}
\subsection{Exponent Distribution}

Both \keyword{float32} and \keyword{bfloat16} use 8 exponent bits,
i.e., they can encode 256 exponent values.
Also both formats have an exponent bias of 127 i.e.,
an exponent of 1 ($2^1$) is stored as $127 + 1 = 128$.
$0$ has an exponent of zero.

The plot below shows the histogram
of the exponent values in one of 
the PaLM-2 layers \cite{model:palm2}.
The X-axis shows the biased exponent value which ranges from $[0, 255]$.
The X2-axis on top shows the corresponding floating point values.
The Y-axis shows the histogram on a $log_{10}$ scale.
The exponent distribution shifts a little but has the same shape for
both weigths and activations.

\begin{figure}[h]
    \centering
    \includegraphics[width=\linewidth]{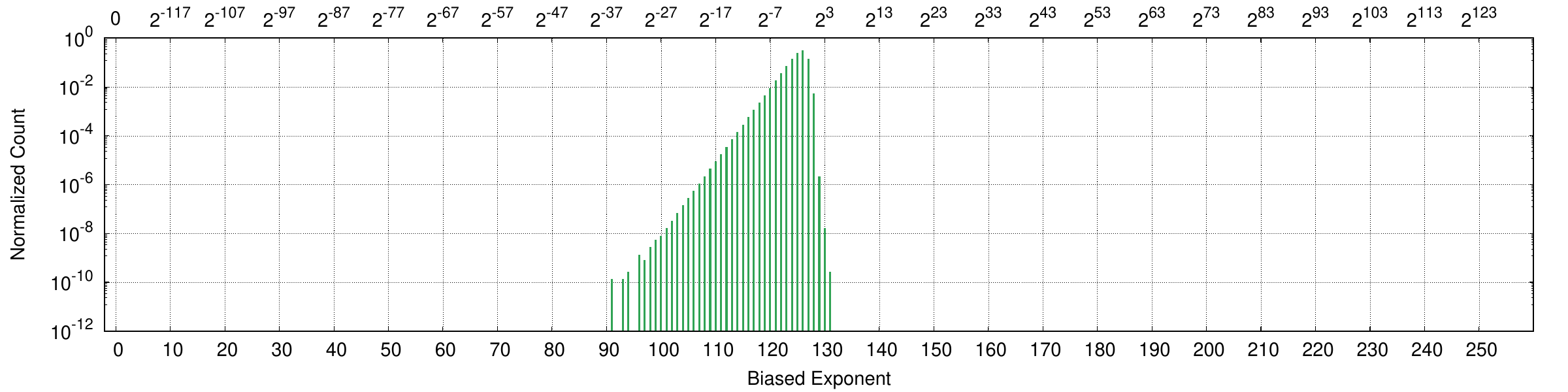}
    \caption{Histogram of the exponent values.}
    \label{fig:exponent_pmf}
\end{figure}

There are multiple observations from this plot:
\begin{itemize}
    \item There are no absolute zeros in the value distribution. However, if the tensor is zero initialized, as is the case for some large embedding
    models, we do observe some absolute zeros.
    \item The left side of the distribution is linear in the log scale.
    For example, the number of values with exponent 101 is two times the
    number of values with exponent 100. The number of values with exponent
    120 is $2^{20} \approx 10^6$, times the number of values with exponent 100.
    This implies that the values are uniformly distributed on the left side.
    \item The distribution reaches a peak and then drops sharply.
    \item The fraction of values with a large magnitude, e.g.
    $[2,16]$ is very small $\approx$ less than 1\%.
    This is because ML models typically, but not always, use weight clipping
    and weight regularization. Models which don't use weight clipping or
    regularization have a higher fraction.
    \item Only a small range, typically $\{0, [80-140]\}$, of biased
    exponents are used. This implies we need only 6 bits, instead of 8 bits,
    for lossless encoding of exponents.
    \item The fraction of values with a small magnitude, e.g.
    $(0, 2^{-10})$ i.e. exponents in the range $[0, 116]$
    is very small $\approx 0.11\%$.
\end{itemize}

The last observation is the most important. In the above distribution, if we
flush values with smaller exponents, i.e. $[0, 116]$ to zero
(exponent of 0) and retain only the top 15 exponents, i.e.
$[117, 131]$, then we need only 4 bits to encode the exponents
$\{0, [117, 131]\}$. The hypothesis is that flushing the exponent tail
to zero will have minimal impact on model quality.

Note that using subnormals and more metadata e.g. \emph{max exponent}
per row, instead of per tensor, significantly reduces the fraction of
values flushed to zero and improves model quality. We found that
\keyword{e4mY} with per tensor metadata and
\keyword{e3mY} with per row or column metadata is quality neutral for
a wide variety of Large Embedding Models (LEMs) and
Large Language Models (LLMs).
\keyword{e2mY} generally requires metadata at finer granularity.
See quality evaluation in Section \ref{sec:quality}.

\subsection{\#Mantissa Bits vs Quality}

Table \ref{tab:palm_2S_accuracy_mantissa} shows the
model quality of the PaLM 2 S model \cite{model:palm2},
for a few LLM datasets as we reduce the number of
mantissa bits of the Feed Forward Networks (FFN) weights,
using Post Training Quantization (PTQ).
The baseline is \keyword{bfloat16}, i.e. \keyword{e8m7}.
We can observe that the model quality is fairly neutral
even with just 1 or 2 mantissa bits.
However, the quality drops significantly with zero
mantissa bits i.e. only power of 2 numbers.

\begin{table}[h]
    \caption{PaLM 2 S quality vs number of mantissa bits.}
    \renewcommand{\arraystretch}{1.5}
    \centering
    \rowcolors{1}{RoyalBlue!10}{}
    \begin{tabular}{l|rrrrrrrr}
\toprule \hline
\rowcolor{RoyalBlue!40}
format     &  base   &  e8m6   &  e8m5   &  e8m4   &   e8m3   &  e8m2   &  e8m1   &  e8m0   \\
hellaswag  &  64.45  &  64.59  &  64.56  &  64.49  &   64.42  &  64.49  &  64.02  &  64.30  \\
lambada    &  84.15  &  84.26  &  84.26  &  83.95  &   84.03  &  83.60  &  82.77  &  65.07  \\
squadv2    &  75.46  &  75.31  &  75.46  &  75.38  &   75.47  &  75.09  &  73.32  &  71.53  \\
triviaqa   &  77.21  &  77.28  &  77.26  &  77.23  &   77.29  &  76.79  &  75.74  &  70.60  \\
webqs      &  23.47  &  23.33  &  23.38  &  23.62  &   23.43  &  23.23  &  21.95  &  20.57  \\
\hline \bottomrule
\end{tabular}
    \label{tab:palm_2S_accuracy_mantissa}
\end{table}

Combining the observations in this and the previous section, we
found that \keyword{e3m1} with per row metadata is fairly quality
neutral for LLMs. \keyword{e2m1} and \keyword{e1m2} benefit from
metadata at finer granularity.
See quality evaluation in Section \ref{sec:quality}.

\section{Applications}

\exmy\ can be used to
(a) Quantize weights, static and dynamic activations and gradients
(b) Quantize master weights and optimizer state
(c) Accelerate compute
(d) Increase multi tenancy
(e) Reduce memory transfers
(f) Reduce network (PCIe, data center network) transfers
(g) Reduce disk storage and disk transfer.

\exmy\ can be used for both Post Training Quantization
(PTQ) and Quantization Aware Training (QAT).
It can be used with both symmetric and affine
quantization schemes.
It can be combined with other techniques e.g. sparsity and
lossless compression algorithms e.g. Zstandard \cite{zstd}.
Since \exmy\ is also a datatype it can be used with other 
quantization recipes and techniques e.g.
HAWQ \cite{papers:hawq-v1}, QLoRA \cite{papers:qlora},
OPTQ \cite{papers:optq}, SmoothQuant \cite{papers:smoothquant} etc.

\exmy\ allows choosing the number of exponent bits, mantissa
bits, and block size on a per tensor basis and hence enables a
gradual trade off between model quality and compression.
The emulation and codecs work on all existing CPUs, GPUs and TPUs,
but can benefit from hardware support for conversions and bit packing and unpacking.

\section{Limitations and Considerations}
\label{sec:limitations}

The \exmy\ datatype itself has no limitations. During serving,
there are no \keyword{NaNs} or \keyword{Infs} and all
encodings have finite values.
During training with \exmy\ emulation, \keyword{NaNs}
and \keyword{Infs} are preserved out of band and
hence all \exmy\ values are still finite.
Training with true \exmy\ encoded values requires
allocating an encoding(s) for these special values
and has not been discussed in this paper.

The \exmy\ technique works best for PTQ of weights
when models use weight regularization, weight clipping etc.,
such that the weights have an exponent distribution as
shown in Fig. \ref{fig:exponent_pmf}.
When those techniques are not used, there is a larger
fraction of values to the right of the peak and that
requires using a format with a bigger dynamic range 
e.g. \keyword{e4m3} instead of \keyword{e3m4}.

Based on the exponent distribution, we can make
educated guesses about the format to use. However, 
the impact on model quality needs to be measured.
Finally, the acceptable change (drop) in model quality
with quantization depends on the trade off between revenue
impact and cost savings.

\section{Quality Evaluation}
\label{sec:quality}

We evaluated \exmy\ on many open source models e.g.
ResNet \cite{model:resnet}, Transformer \cite{model:transformer},
BERT \cite{model:bert}, as well as many internal
vision, ranking, recommendation,
Large Embedding Models (LEMs) and Large Language Models (LLMs).
In this section, we show the quality evaluation on the PaLM 2 S
model \cite{model:palm2} using the following datasets:
Adversarial NLI (ANLI) \cite{dataset:anli},
ARC \cite{dataset:arc},
BoolQ \cite{dataset:boolq},
CB,
COPA \cite{dataset:copa},
COQA,
DROP \cite{dataset:drop},
HellaSwag \cite{dataset:hellaswag},
LAMBADA \cite{dataset:lambada},
Natural Questions \cite{dataset:natural_questions},
OpenBookQA \cite{dataset:openbookqa},
PIQA \cite{dataset:piqa},
QuAC \cite{dataset:quac},
RACE \cite{dataset:race},
ReCoRD \cite{dataset:record},
RTE,
SQuAD v2 \cite{dataset:squadv2},
StoryCloze \cite{dataset:storycloze},
TriviaQA \cite{dataset:triviaqa},
TyDi QA \cite{dataset:tydiqa},
WebQuestions \cite{dataset:webqs},
WiC \cite{dataset:wic},
Winograd \cite{dataset:wsc}, and
WinoGrande \cite{dataset:winogrande}.

The left half of Table \ref{tab:palm_2S_accuracy_format},
shows the scores 
when all the Feed Forward Network (FFN) weights are post training
quantized to \keyword{e3m4}, \keyword{e3m3}, \keyword{e3m2},
\keyword{e3m1}, \keyword{e3m0}, and \keyword{e2m1} respectively.
The attention layers are always quantized to \keyword{e3m4}.
The block size is the length of the row in the weight matrix.
The scheme is maximum exponent before rounding.
There are multiple observations from the table:

\begin{table}[!h]
    \caption{PaLM 2 S quality at different \exmy\ formats and block sizes.}
    \renewcommand{\arraystretch}{1.25}
    \centering
    \rowcolors{1}{RoyalBlue!10}{}
    \begin{tabular}{l|rrrrrrr|rrrr}
\toprule \hline

\rowcolor{RoyalBlue!40}
{format}                       &  base   &  e3m4   &  e3m3   &  e3m2   &  e3m1   &  e3m0   &  e2m1   &  e2m1         &  e2m1         &  e2m1         &  e2m1        \\

\rowcolor{RoyalBlue!40}
block\_size                    &  {}     & {row}   &  {row}  &  {row}  &  {row}  &  {row}  &  {row}  &  512          & 256           & 128           & 64           \\

\midrule
anlir1                         &  53.00  &  53.90  &  54.10  &  53.80  &  54.30  &  52.40  &  51.30  &  52.90        &  53.80        &  53.70        &  53.70       \\
anlir2                         &  49.00  &  49.30  &  49.10  &  48.60  &  49.30  &  46.90  &  47.20  &  45.40        &  46.80        &  47.60        &  48.20       \\
anlir3                         &  52.75  &  53.08  &  53.42  &  53.17  &  53.92  &  53.92  &  52.08  &  53.00        &  52.50        &  51.75        &  52.75       \\
arcchallenge                   &  56.06  &  56.57  &  56.74  &  56.74  &  55.46  &  54.69  &  52.13  &  54.86        &  55.03        &  55.63        &  55.63       \\
arceasy                        &  84.93  &  84.89  &  84.89  &  85.06  &  84.05  &  84.13  &  78.96  &  82.74        &  82.87        &  83.21        &  83.54       \\
boolq                          &  89.08  &  88.81  &  88.93  &  88.90  &  88.96  &  86.88  &  79.08  &  85.57        &  87.80        &  88.29        &  88.13       \\
cb                             &  87.50  &  87.50  &  85.71  &  91.07  &  83.93  &  87.50  &  76.79  &  82.14        &  83.93        &  85.71        &  85.71       \\
copa                           &  89.00  &  88.00  &  87.00  &  87.00  &  89.00  &  88.00  &  89.00  &  86.00        &  87.00        &  87.00        &  88.00       \\
coqa                           &  63.06  &  63.21  &  63.22  &  62.92  &  62.81  &  59.64  &  61.44  &  62.73        &  62.51        &  62.29        &  62.73       \\
drop                           &  54.60  &  54.64  &  54.56  &  54.24  &  53.87  &  50.06  &  51.68  &  53.32        &  53.33        &  53.75        &  53.97       \\
hellaswag                      &  64.45  &  64.53  &  64.33  &  64.54  &  64.01  &  64.20  &  62.32  &  64.02        &  63.68        &  63.68        &  63.13       \\
lambada                        &  84.15  &  84.30  &  84.34  &  83.58  &  83.27  &  65.07  &  75.57  &  79.64        &  80.38        &  82.50        &  83.17       \\
eue19\_defr                    &  36.17  &  35.96  &  35.91  &  35.58  &  35.71  &  34.17  &  31.62  &  34.98        &  35.10        &  35.55        &  35.45       \\
eue19\_frde                    &  26.79  &  26.66  &  26.09  &  24.93  &  27.31  &  26.62  &  21.04  &  25.78        &  25.94        &  26.11        &  26.53       \\
wmt14\_enfr                    &  44.89  &  44.96  &  45.07  &  45.06  &  44.44  &  41.79  &  41.42  &  43.35        &  43.35        &  43.69        &  43.97       \\
wmt14\_fren                    &  44.96  &  44.85  &  45.26  &  44.80  &  44.74  &  41.29  &  41.56  &  43.92        &  44.41        &  44.56        &  44.48       \\
wmt16\_deen                    &  48.37  &  48.56  &  48.53  &  48.29  &  48.02  &  45.24  &  44.59  &  47.70        &  47.82        &  48.10        &  47.90       \\
wmt16\_ende                    &  39.44  &  39.37  &  39.32  &  39.13  &  38.75  &  34.20  &  35.65  &  37.73        &  38.16        &  37.82        &  38.42       \\
wmt16\_enro                    &  32.63  &  32.66  &  32.72  &  32.50  &  32.76  &  31.70  &  31.53  &  32.46        &  32.46        &  32.72        &  32.68       \\
wmt16\_roen                    &  46.62  &  46.59  &  46.50  &  46.62  &  46.27  &  44.84  &  44.08  &  45.35        &  45.96        &  45.87        &  45.92       \\
wmt19\_enkk                    &  8.36   &  8.48   &  8.24   &  8.66   &  8.53   &  5.82   &  7.71   &  7.88         &  7.09         &  7.01         &  7.61        \\
wmt19\_enzh                    &  5.28   &  5.12   &  5.20   &  5.06   &  4.99   &  4.46   &  5.90   &  5.38         &  5.48         &  4.86         &  4.87        \\
wmt19\_kken                    &  31.15  &  31.23  &  31.41  &  31.16  &  31.07  &  28.99  &  26.46  &  29.97        &  30.32        &  30.71        &  30.92       \\
wmt19\_zhen                    &  32.76  &  32.63  &  32.47  &  32.43  &  31.57  &  28.12  &  28.43  &  30.56        &  30.88        &  30.81        &  31.54       \\
nqs                            &  27.92  &  28.06  &  27.78  &  27.29  &  26.32  &  22.35  &  20.00  &  24.71        &  24.85        &  24.46        &  26.04       \\
openbookqa                     &  47.80  &  47.60  &  47.80  &  48.40  &  46.40  &  44.60  &  42.40  &  47.00        &  46.20        &  45.60        &  47.20       \\
piqa                           &  81.01  &  81.18  &  81.18  &  81.23  &  80.85  &  80.36  &  77.97  &  80.79        &  80.69        &  80.63        &  80.85       \\
quac                           &  23.46  &  23.43  &  23.42  &  23.39  &  22.83  &  19.87  &  20.51  &  22.49        &  22.80        &  22.70        &  22.57       \\
raceh                          &  48.31  &  48.28  &  48.68  &  48.68  &  48.74  &  48.48  &  47.20  &  49.06        &  48.54        &  48.91        &  48.80       \\
racem                          &  64.83  &  64.97  &  65.67  &  65.04  &  65.04  &  63.79  &  63.02  &  64.48        &  64.55        &  64.76        &  64.69       \\
record                         &  92.10  &  92.22  &  92.02  &  92.15  &  91.93  &  89.44  &  89.34  &  91.20        &  91.37        &  91.73        &  91.80       \\
rte                            &  77.26  &  77.62  &  77.26  &  78.34  &  77.98  &  75.45  &  79.42  &  77.98        &  77.62        &  77.98        &  77.98       \\
squadv2                        &  75.46  &  75.63  &  75.63  &  75.25  &  73.52  &  71.70  &  75.41  &  76.18        &  74.98        &  75.19        &  74.73       \\
storycloze                     &  81.88  &  81.83  &  82.36  &  82.26  &  81.51  &  82.31  &  78.67  &  81.93        &  81.56        &  81.56        &  81.13       \\
triviaqa                       &  77.21  &  77.33  &  77.43  &  76.77  &  75.82  &  71.01  &  67.43  &  73.90        &  74.30        &  74.68        &  75.77       \\
tydiaqa                        &  17.31  &  17.33  &  17.14  &  17.02  &  16.47  &  14.24  &  14.08  &  16.27        &  16.15        &  16.15        &  16.09       \\
webqs                          &  23.47  &  23.47  &  23.18  &  23.03  &  21.65  &  20.62  &  17.27  &  22.24        &  21.21        &  21.46        &  22.24       \\
wic                            &  51.10  &  51.25  &  50.47  &  53.61  &  50.94  &  50.31  &  50.16  &  50.00        &  50.31        &  50.63        &  50.16       \\
winograd                       &  84.98  &  84.98  &  85.71  &  84.25  &  84.25  &  82.78  &  79.12  &  84.62        &  82.78        &  84.98        &  83.15       \\
winogrande                     &  77.35  &  77.82  &  77.03  &  78.14  &  77.19  &  75.22  &  69.46  &  73.40        &  74.11        &  76.40        &  75.37       \\
wsc273                         &  84.62  &  85.35  &  84.98  &  83.88  &  84.62  &  82.05  &  77.66  &  83.15        &  82.05        &  84.98        &  81.32       \\
\hline \bottomrule
\end{tabular}
    \label{tab:palm_2S_accuracy_format}
\end{table}

\begin{itemize}

\item Overall, LLMs hold their quality very well with simple 
PTQ of weights down to \keyword{e3m1} even with per-row
metadata and without requiring any advanced techniques like
SmoothQuant \cite{papers:smoothquant}, OPTQ \cite{papers:optq},
ZeroQuant \cite{papers:zeroquant} etc.

\item The quality does not decrease
monotonically as we reduce the number of exponent and/or
mantissa bits. For example, for OpenBookQA and PIQA,
the quality with \keyword{e3m2} is
better than \keyword{bfloat16}, which is \keyword{e8m7}.
We suspect this is due to the opposing effects of
quantization and regularization.

\item Some datasets e.g. HellaSwag are very resilient to
quantization, while others e.g. LAMBADA are more sensitive,
i.e. the choice of the quantization format is dataset dependent.

\item The quality drop is significant at 4 bit formats e.g.
\keyword{e3m0} and \keyword{e2m1} at large block sizes.

\end{itemize}

The quality of \keyword{e2m1} improves by reducing
the block size. The right half of
Table \ref{tab:palm_2S_accuracy_format}, shows the scores
when the block size is reduced from \keyword{row} to 512 and
then to 64 in powers of 2.
We can observe that for sensitive datasets like LAMBADA, the 
quality increases monotonically as we decrease the block size.

\section{Related Work}


Posits \cite{format:posit, papers:posit_ai} are an
alternative way of representing real numbers. They
offer a good trade-off between dynamic range and accuracy,
encounter fewer exceptions, and have tapered precision
i.e. numbers near $\pm$1 have more precision,
while very big and very small numbers have less.
Other floating point formats have also been proposed e.g.
Logarithmic numbers \cite{format:log} and
NormalFloat4 \cite{papers:qlora} which targets 
normally distributed weights.

Numerous studies and techniques compare and use different data types in various settings, such as post training quantization (PTQ),
quantization aware training (QAT) and fully quantized training (FQT).
For quantized inference, multiple industry and academic white papers
highlight the overall benefits and general approaches to
\keyword{int8} (sometimes even \keyword{int4}) quantization
\cite{papers:nvidia_integer,
papers:nagel2021white, gholami2022survey} exploring  quantization
granularity, scaling methods, initialization techniques, and data
formats.

For LLM quantization, a plethora of techniques have emerged such as
one-shot PTQ techniques with layer-wise
optimizations~\cite{papers:optq}, optimization free techniques
which leverage robustness of data types
(\keyword{fp8})~\cite{papers:power_of_exponent}, and 4 bit techniques with
searches for exponents bits and clipping range~\cite{liu2023llm}.
After analyzing the causes of quality degradation in
LLM quantization, various authors have identified outlier
behaviour to be
problematic and proposed various solutions, such as offline
transformation of weights to absorb
outliers~\cite{papers:smoothquant}, channel-wise shifting and
scaling~\cite{wei2023outlier}, rotation of hidden state
matrices~\cite{ashkboos2024quarot}, modifications of the attention
mechanism~\cite{bondarenko2024quantizable}, and mixed-precision matrix
decomposition~\cite{dettmers2022gpt3}.

To combat model quality degradation at lower bit-widths, some
previous
works propose mixed precision approaches which keep sensitive layers
in higher precision, whereby the sensitivity is usually approximated
through a Hessian ~\cite{papers:hawq-v1, papers:hawq-v2,
papers:hawq-v3,schaefer2023augmenting}. Alternatively, to improve
quality other works incorporate quantization consideration into
training (QAT), for example through optimizing clipping
scalars~\cite{sakr2022optimal} or data-free distillation method based
on outputs of a pretrained model~\cite{liu2023llm}.

Extending quantization to training (FQT), QLoRA~\cite{papers:qlora}
reduces the memory requirements for LLM finetuning by quantizing the
weights of the frozen pretrained model to 4 bits. Going even
further~\cite{mellempudi2019mixed} quantize weights, activations,
errors, gradients, and the master copy of the weights during training
and achieve SOTA through various data sets and models utilizing loss
scaling method to augment the reduced subnormal range and stochastic
rounding. Attempting to simplify training with
FP8~\cite{blake2023unit} present unit scaling, a paradigm which
yields
unit variance for weights, activations, and gradients at
initialization. This approach works without quality degradation
across multiple optimizers and models.

\section{Conclusion}

In this work, we described \exmy, a new data type and
technique for quantization of ML models.
It can represent arbitrary bit width signed integers,
symmetric signed integers and floating point numbers.
It supports subnormals and arbitrary block shapes and
sizes.

We described a novel bit packing scheme which achieves
perfect compression using existing storage data types.
It works for all arbitrary bit widths and is amenable to
vector processing on all existing hardware.
The scheme offers byte addressability and works
seamlessly with array sharding.
We implemented libraries for emulation, encoding and
decoding tensors in multiple frameworks.

We discovered the distribution of exponents in ML models.
We described a technique to exploit it and significantly
reduce the number of bits required by the model while 
retaining model quality.
The technique can be used to quantize master weights,
training weights, serving weights, static and dynamic
activations, gradients and network communication.
This reduces CPU RAM footprint and bandwidth, accelerator
RAM (HBM) footprint and bandwidth, PCIe and network
latency, disk I/O and increases multi-tenancy.
With hardware support the technique can also be used for
compute acceleration.

\exmy\ has been deployed in production by multiple teams.
We have found many interesting applications and
hope the community at large will embrace arbitrary bit
widths and formats to develop novel techniques
and applications.

\begin{ack}
Abdullah Rashwan,
Afroz Mohiuddin,
Alex Tomala,
Andy Chu,
Anselm Levskaya,
Bing-Rong Lin,
Chen Chen,
Chris Waterson,
Clemens Schaefer,
Dar-Shyang Lee,
David Majnemer,
Diego Caballero,
Eric Chang,
Erica Moreira,
Eugene Zhulenev,
Grant Wang,
Gil Tabak,
Guangda Lai,
Jacques Liao,
Jaideep Singh,
Jayant Madhavan,
Jomy Alappattu,
Jon Clark,
Kirill Borozdin,
Kirk Sanders,
Lili Hu,
Marissa Ikonomidis,
Matthew Fahrbach,
Michael Mangan,
Ming Liu,
Nand Dalal,
Naveen Kumar,
Navid Lambert-Shirzad,
Nathan Lintz,
Orhan Firat,
Pierre-François Laquerre,
Pidong Wang,
Phuong Dao,
Pooja Aggarwal,
Qi Lyu,
Qi Wu,
Rahul Nagarajan,
Rajkumar Samuel,
Rakshith Reddy Polireddy,
Reza Sadoddin,
Rigel Swavely,
Robin Sabhnani,
Rohan Anil,
Ryan Doherty,
Sencer Selcuk,
Shaolin Qu,
Shen Wu,
Shicheng Xu,
Shruti Gupta,
Srikanth Dwarakanath,
Tao Yu,
Tom Jablin,
Victor Akabutu,
Woohyun Han,
Yang Li,
Yiwen Deng,
Yuan Huang,
Zongwei Zhou.
\end{ack}

\bibliography{citations/datasets, citations/formats, citations/software, citations/models, citations/misc, citations/papers, citations/hardware}

\end{document}